\pdfoutput=1

\documentclass[11pt]{article}

\usepackage[]{acl}

\usepackage{times}
\usepackage{latexsym}

\usepackage[T1]{fontenc}

\usepackage[utf8]{inputenc}

\usepackage{microtype}

\usepackage{inconsolata}

\usepackage{graphicx} 
\usepackage{subcaption}  
\usepackage{kotex} 
\usepackage{amssymb}
\usepackage{amsmath}
\usepackage{sansmath}
\usepackage{xcolor}
\usepackage{booktabs}
\usepackage{multirow}
\usepackage{multicol}
\usepackage{bm}

\title{Unveiling Key Aspects of Fine-Tuning in Sentence Embeddings:\\A Representation Rank Analysis}

\author{Euna Jung$^{1*\dag}$, Jaeill Kim$^2$\thanks{Equal contribution.}$^{\dag}$, Jungmin Ko$^3$, Jinwoo Park$^{1\dag}$, Wonjong Rhee$^{3,4\ddag}$\\
$^1$Samsung Advanced Institute of Technology\hspace{5mm}$^2$LINE Investment Technologies\\
$^3$Interdisciplinary Program in Artificial Intelligence, Seoul National University\\
$^4$Department of Intelligence and Information \& RICS, Seoul National University\\
{\small $^1$\texttt{{\{una.jung,jw0109.park\}}@samsung.com}}\hspace{3mm}{\small $^2$\texttt{jaeill.kim@linecorp.com}}\hspace{3mm}
{\small $^3$\texttt{\{jungminko,wrhee\}@snu.ac.kr}}}

\begin{document}
\maketitle
\def\thefootnote{\dag}\footnotetext{Work performed while at Seoul National University.}
\def\thefootnote{\ddag}\footnotetext{Corresponding author.}

\begin{abstract}
The latest advancements in unsupervised learning of sentence embeddings predominantly involve employing contrastive learning-based (CL-based) fine-tuning over pre-trained language models. In this study, we analyze the latest sentence embedding methods by adopting representation rank as the primary tool of analysis. We first define Phase 1 and Phase 2 of fine-tuning based on when representation rank peaks. Utilizing these phases, we conduct a thorough analysis and obtain essential findings across key aspects, including alignment and uniformity, linguistic abilities, and correlation between performance and rank. For instance, we find that the dynamics of the key aspects can undergo significant changes as fine-tuning transitions from Phase 1 to Phase 2. Based on these findings, we experiment with a rank reduction (RR) strategy that facilitates rapid and stable fine-tuning of the latest CL-based methods. Through empirical investigations, we showcase the efficacy of RR in enhancing the performance and stability of five state-of-the-art sentence embedding methods.
\end{abstract}
\section{Introduction}
\label{sec:introduction}

\begin{figure}[!t]
    \centering
    \includegraphics[width=\linewidth]{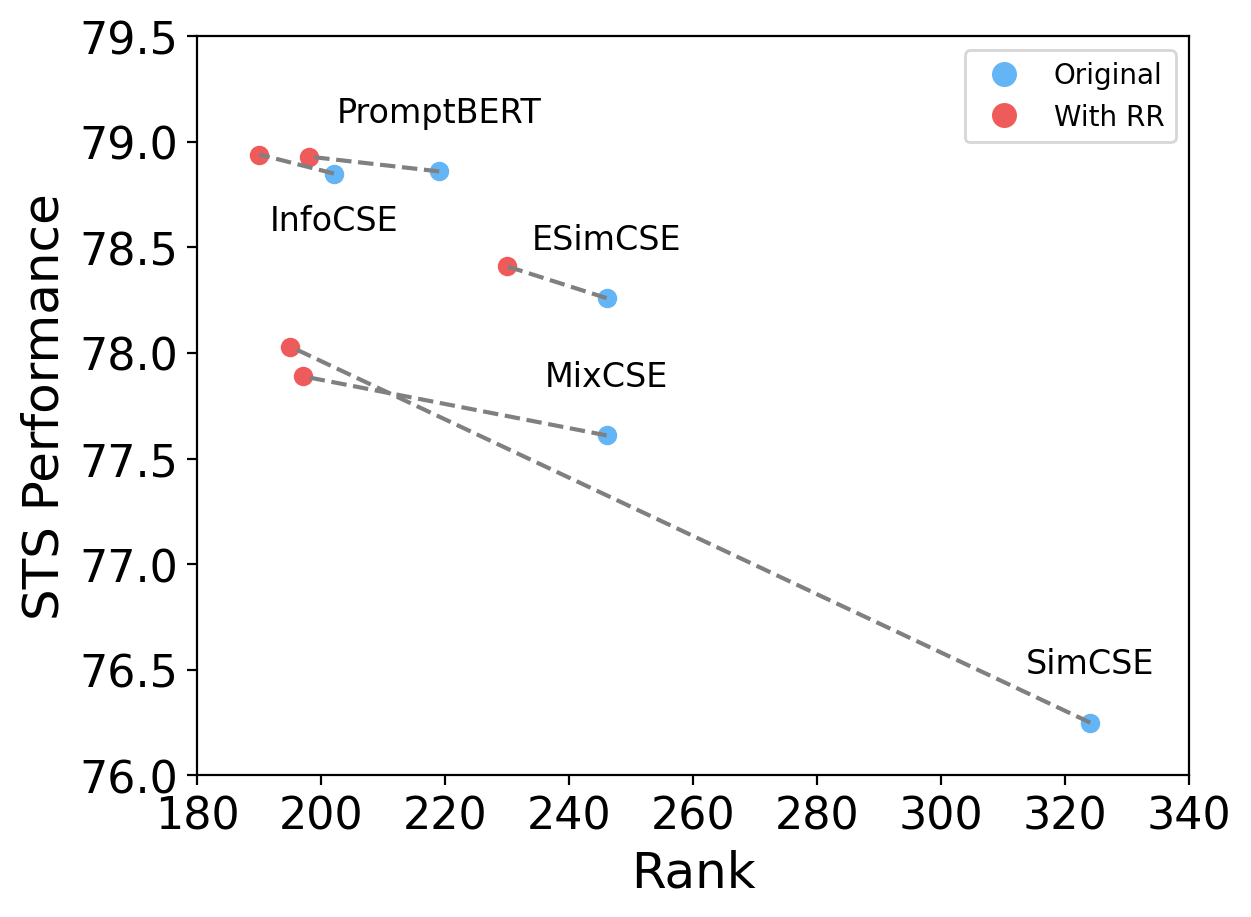}
    \caption{Representation rank and Semantic Textual Similarity~(STS) performance of various CL-based models. Blue dots depict the original models and the red dots represent the same models fine-tuned with Rank Reduction (RR) regularizer.}
    \label{fig:rank_performance}
\end{figure}

Learning sentence embeddings, which refer to the vectorized representations reflecting the semanticity of given textual inputs, is an essential task in the field of natural language processing (NLP).
Following the outstanding performance of SimCSE~\citep{gao2021simcse}, which employed contrastive learning as the unsupervised fine-tuning method for the sentence embedding task, Contrastive Learning-based (CL-based) models have emerged as predominant approaches in the field of sentence embedding~\citep{chuang2022diffcse,jiang2022improved,wu2022pcl,zeng2022contrastive,wu2021esimcse,zhang2022unsupervised,jiang2022promptbert,klein2022micse}.

Contrastive learning has showcased remarkable performance across various domains, including visual self-supervised learning~\citep{chen2020simple} and multi-modal learning~\citep{radford2021learning}. Consequently, numerous studies have focused on deepening our understanding of contrastive learning. However, contrastive learning for sentence embedding is distinct in that it is employed as a fine-tuning method on a pre-trained language model. A pre-trained language model already possesses a range of natural language processing capabilities~\citep{devlin2018bert,hewitt2019structural}. Therefore, fine-tuning must ensure the preservation of linguistic abilities beneficial for sentence embedding while simultaneously reducing the contrastive loss, which directly impacts the Semantic Textual Similarity~(STS) task performance. When fine-tuning with contrastive loss, the embeddings must be adequately represented to enable cosine similarity, the predominant metric for utilizing and evaluating semantic embeddings, to serve effectively as the primary tool for semantic analysis.
Despite these distinctions, most of the existing studies have primarily focused on enhancing STS performance and have provided only limited analysis directly related to their proposed methods.

Recently, there has been a notable emphasis on utilizing representation rank for the analysis of deep learning models~\citep{zhuo2023towards,garrido2023rankme,Kim_2023_CVPR}. Inspired by these recent works, we conduct an in-depth analysis of CL-based fine-tuning by thoroughly investigating representation rank and several key aspects closely related to it. Specifically, we examine uniformity and alignment, linguistic abilities, and the correlation between STS performance and rank. All these key aspects are analyzed in conjunction with Phase 1 and Phase 2 of fine-tuning, which are defined based on when representation rank peaks during the fine-tuning process. We observe and report a strong linkage with representation rank for all the key aspects.

Motivated by these strong findings, we conduct experiments incorporating a rank regularization approach, specifically \textit{Rank Reduction~(RR)} regularizer, during the training of CL-based models. The experimental results highlight the effectiveness of RR in reducing rank, leading to increased sentence embedding performance and enhanced stability. Indeed, various CL-based models have demonstrated a trend towards decreasing rank and increasing performance in their evolution, as illustrated in Figure~\ref{fig:rank_performance}.

This study enhances our understanding of contrastive learning-based fine-tuning of language models through representation rank analysis. Our analysis relies on defining two distinct phases of fine-tuning process, wherein key aspects can be affected in opposite manners in the two phases. We demonstrate adjusting rank is effective for performance enhancement and straightforward to implement.

\section{Background and Related Works}

\subsection{Contrastive Learning-Based Models}
\subsubsection{SimCSE Training through Contrastive Self-Supervised Learning}
Unsupervised SimCSE training leverages contrastive SSL, where embeddings of a positive pair are pulled together while those from negative pairs are pushed apart.
Following the framework of \citet{chen2020simple}, the loss is defined as 
\begin{align}
\mathcal{L}_\mathsf{SimCSE}= - \underset{i}{\mathbb{E}} \left[\log \frac{\mathrm{exp}(\bm{h}_i^T \bm{h}'_i / \tau)}{\sum_{j=1}^N \mathrm{exp}(\bm{h}_i^T \bm{h}'_j / \tau)}\right],
\label{eq:ce_loss}
\end{align}
where the two representations of input $\bm{x}_i$ under different dropout masks are denoted as $\bm{h}_i$ and $\bm{h}'_i$.

\subsubsection{Models Adopting the Framework of SimCSE}
Following SimCSE, numerous subsequent studies adopting the SimCSE framework have emerged, resulting in the proposal of various CL-based models. Among these, we selected four models based on performance and reproducibility, in addition to SimCSE itself.
We use these CL-based models to analyze the relationship between representational rank and sentence embedding performance.
In the following paragraphs, we provide brief descriptions of each model.

\textbf{MixCSE}~\citep{zhang2022unsupervised} enhances sentence embedding performance by creating hard negatives by mixing positive and negative features. The authors emphasize the crucial role of hard negatives in contrastive learning for maintaining a robust gradient signal.

\textbf{ESimCSE}~\citep{wu2021esimcse} addresses length bias between positive sentence pairs by repeating parts of the input text and employing momentum contrast to increase the number of negative pairs. Through these strategies, ESimCSE effectively mitigates length bias while improving sentence embedding performance.

\textbf{InfoCSE}~\citep{wu2022infocse} introduces an additional masked language model architecture to make the [CLS] representation of the model aggregate more information. This model, utilizing an auxiliary objective, demonstrates effectiveness and transferability in sentence embedding tasks.

\textbf{PromptBERT}~\citep{jiang2022promptbert} significantly boosts the performance of sentence embedding by employing a prompt and template denoising technique. It stands out as one of the state-of-the-art models. The performance improvement in PromptBERT is attributed to both finding the optimal prompt and removing bias introduced by the template. In our study, we analyze the impact of these two approaches on the rank of representations.

Among recent studies, MixCSE and PromptBERT have highlighted the presence of an instability problem in learning unsupervised SimCSE, where the performance of SimCSE exhibits significant variance depending on the random seed. We deal with this instability problem as well in this study.

\subsection{Representation Rank}
\subsubsection{Measuring and Regularizing Representation Rank}
\label{subsubsec:measuring_rank}
Conventionally, the representation rank is determined by counting the number of largest singular values that capture a substantial portion of total singular value energy in the representation matrix $\bm{H} \in \mathbb{R}^{N \times d}$, where $\bm{H}(\triangleq[\bm{h}_1, ..., \bm{h}_N]^T)$ denotes a stack of representations $\bm{h}_i \in \mathbb{R}^{d}$. The variables $N$ and $d$ represent the batch size and the dimension of the representation, respectively. Throughout this study, we employ this energy-based measurement for the quantification of rank.

While the energy-based rank is intuitive and useful for analysis, the discrete nature of rank measurement presents challenges in its direct application for regularization purposes.
To address this challenge, we adopt the effective rank~\citep{roy2007effective} as an approximation of the rank, defined as 
\begin{align}
\mathsf{erank}(\bm{H}) \triangleq \exp (-\sum_j \lambda_{j} \log{\lambda_{j}}), \label{eq:approx_rank}
\end{align}
where $\{ \lambda_{j} \}$ are the eigenvalues of $\bm{Z}^T\bm{Z} / N$, $\bm{Z}\triangleq[\bm{z}_1, ..., \bm{z}_N]^T$, and $\bm{z}_i \triangleq \bm{h}_i / \lVert \bm{h}_i \rVert_2$.
Note that $\bm{Z}^T\bm{Z} / N\ge0$ and $tr(\bm{Z}^T\bm{Z} / N) = 1$.
In our study, we utilize the logarithm of Eq.~\ref{eq:approx_rank} for the purpose of training.
This choice is motivated by the efficacy demonstrated in~\citet{Kim_2023_CVPR}.

\subsubsection{Rank and Contrastive Learning}
\label{subsec: rank_contrastive_learning}
Recently, the \textit{rank} of representation has been studied for many deep learning research topics, as it directly reflects the dimensionality utilized by the representation.
In~\citet{jing2021understanding}, it has been observed that contrastive learning is susceptible to the phenomenon of \textit{dimensional collapse}, wherein representation vectors tend to concentrate within a lower-dimensional subspace rather than spanning the entirety of the available embedding space.
In~\citet{garrido2023rankme}, rank of representation is identified as a robust predictor of the downstream performance.
For self-supervised learning, investigations such as those presented in~\citet{hua2021feature} and \citet{Kim_2023_CVPR} have presented findings indicating that increasing rank of a representation can contribute to the enhancement of performance in self-supervised learning.
In the broader research domain of unsupervised representation learning, \citet{zhuo2023towards} have revealed a phenomenon wherein the training dynamics of rank for pre-training vision tasks experience a rapid decrease followed by an increase. 
In contrast, we have found an opposite trend for CL-based fine-tuning, as we will elaborate in Section~\ref{sec:SimCSEandRank}.
Research on rank, as described, has predominantly occurred within the vision field and has been confined to models trained from scratch. In this study, we study rank of CL-based language models trained through fine-tuning.

\section{Key Aspects of Fine-Tuning}  
\label{sec:SimCSEandRank}

\begin{figure}[!t]
    \centering
    \includegraphics[width=0.9\linewidth]{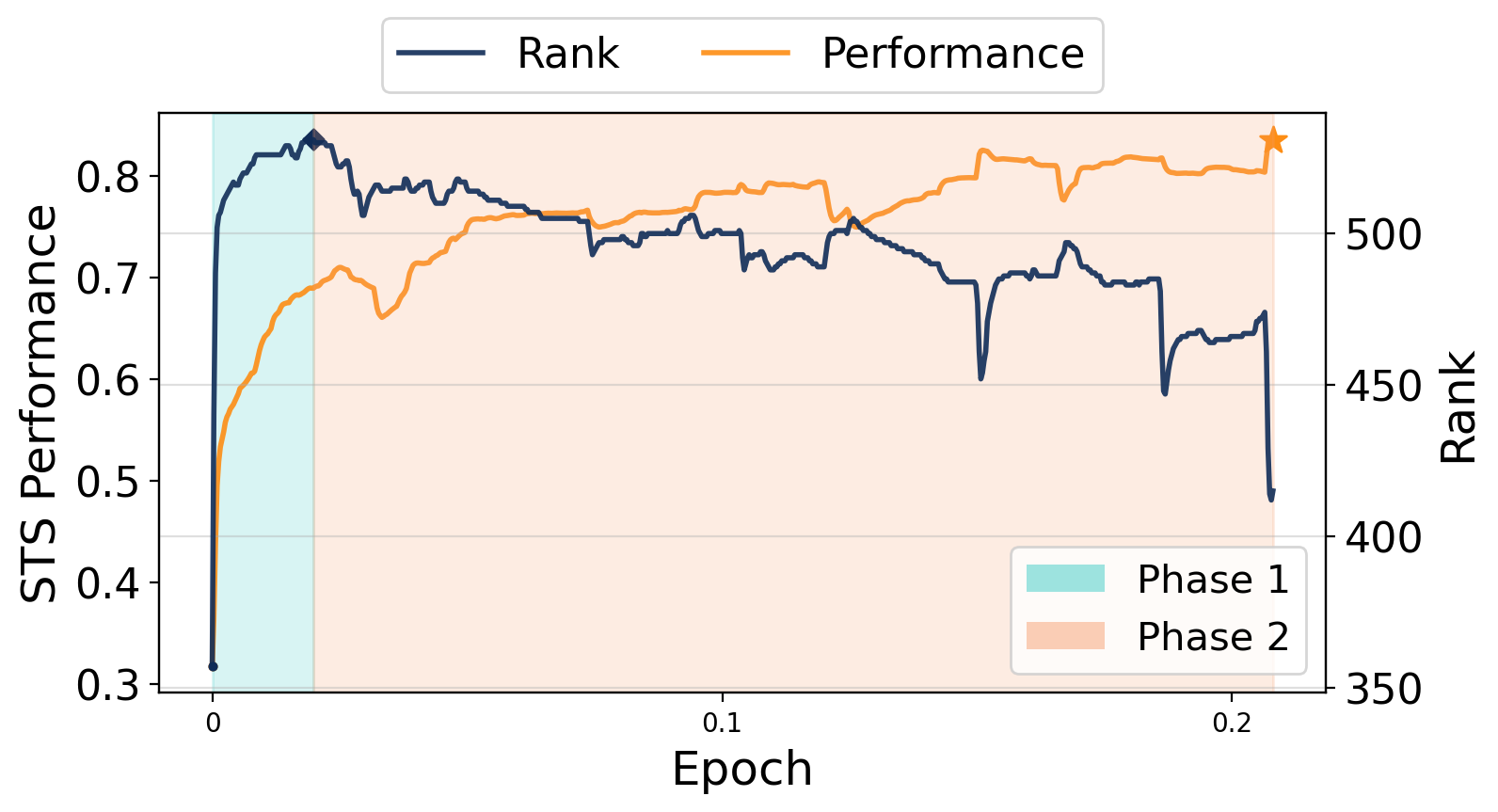}
    \caption{Training dynamics for fine-tuning BERT with contrastive learning. Phase 1 and Phase 2 represent two distinct stages of fine-tuning, delineated by the peak in representation rank.}
    \label{fig:phase_definition}
\end{figure}

In this section, we analyze key aspects of fine-tuning language model in sentence embeddings. Specifically, we consider SimCSE where BERT is fine-tuned using contrastive learning. Other CL-based sentence embedding methods will be considered in Section~\ref{sec:Experiments}. We commence our analysis by visualizing fine-tuning dynamics of validation performance and representation rank, as depicted in Figure~\ref{fig:phase_definition}. While the validation performance tends to improve with fine-tuning, representation rank increases sharply initially and then begins to decline as the fine-tuning progresses. 
The fine-tuning dynamics can be evidently divided into two phases based on rank. We define \textit{Phase 1} as the initial phase that spans from the onset of training to the point where the rank reaches its maximum value. The subsequent phase, defined as \textit{Phase 2}, covers the period extending beyond the conclusion of Phase 1 until the model achieves its best validation performance. It is noted that the final model is selected at the conclusion of Phase 2, commonly referred to as early stopping, even though the training continues until the pre-determined length of one epoch.
Based on the two distinct phases, we explore the key aspects of fine-tuning in the following.

\subsection{Alignment and Uniformity}
\label{subsec:alignment_uniformity}

\begin{figure}[!t]
    \centering
    \includegraphics[width=0.9\linewidth]{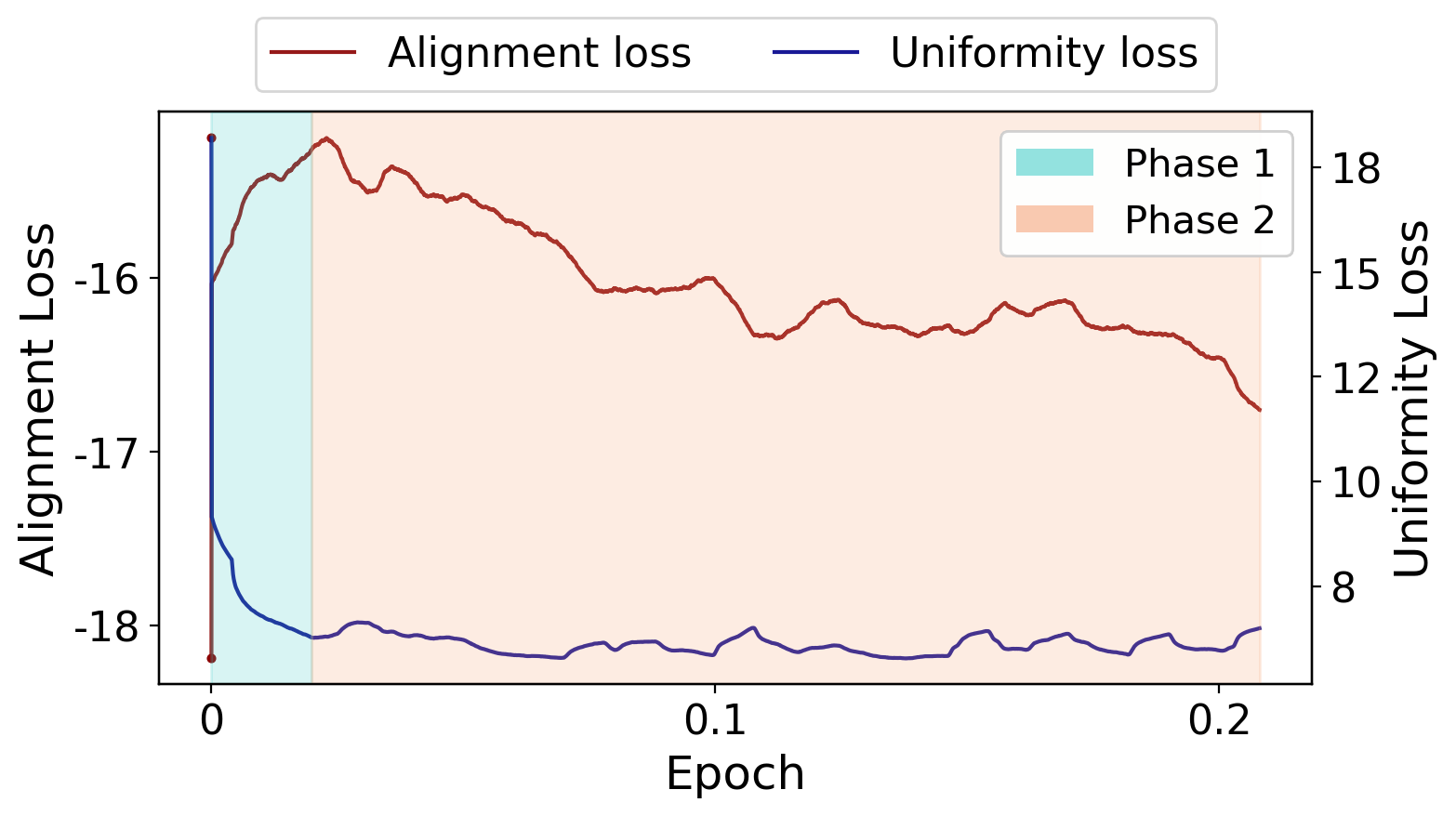}
    \caption{Training dynamics of alignment loss and uniformity loss. Initially, alignment loss starts low near the bottom of the y-axis, while uniformity loss begins high near the top. However, both experience a sharp change at the outset, resulting in a significant overlap between the curves.}
    \label{fig:uniformity_alignment}
\end{figure}

Contrastive loss optimizes two distinct properties of representation~\citep{wang2020understanding}: alignment and uniformity. To quantify these properties, we employ the following formula introduced in Decoupled Contrastive Learning (DCL)~\citep{yeh2022decoupled}, which has been also utilized for the training dynamics analysis of sentence embeddings in a recent work~\citep{nie2023inadequacy}.

\begin{align}
\mathcal{L}_\mathsf{dcl} \triangleq& - \underset{i}{\mathbb{E}} \left[\log \frac{\mathrm{exp}(h_i^T h'_i / \tau )}{\sum_{j=1, j\ne i}^N\mathrm{exp}(h_i^T h'_j / \tau )}\right] \label{eq:dcl_loss}\\
=&\underbrace{-\underset{i}{\mathbb{E}}\left[h_i^T h'_i / \tau\right]}_{\mathsf{alignment\ loss}} \label{eq:pos_loss}\\
&+ \underbrace{\underset{i}{\mathbb{E}}\left[\log \sum_{j=1, j\ne i}^N\mathrm{exp}(h_i^T h'_j / \tau)\right]}_{\mathsf{uniformity\ loss}}. \label{eq:neg_loss}
\end{align}

In Figure~\ref{fig:uniformity_alignment}, a significant improvement in uniformity is observed during Phase 1, followed by a stabilizing trend in Phase 2. Conversely, there is a significant deterioration in alignment during Phase 1, succeeded by a moderate improvement trend in Phase 2. This observation makes it clear that the contrastive loss prioritizes enhancing uniformity over alignment during Phase 1. It suggests that BERT embeddings are initially far from being uniformly distributed and fine-tuning focuses on improving uniformity such that cosine similarity measure, that is used as the metric of sentence embedding, can function effectively. Once uniformity is sufficiently improved in Phase 1, fine-tuning transitions to Phase 2 where it focuses on recovering alignment while maintaining uniformity.

Rank is related to uniformity. For example, when autocorrelation matrix of representation, $\bm{Z}$, is an identity matrix, it is trivial to show that both representation rank and uniformity are maximized. 
Also, it can be observed in Phase 1 that the rank in Figure~\ref{fig:phase_definition} and the uniformity loss in Figure~\ref{fig:uniformity_alignment} undergo steep adjustments with almost reversed shapes. This further emphasizes the strong relationship between rank and uniformity. Rank is also closely related to alignment. In Phase 2, rank exhibits a very similar trend to alignment, with both steadily returning towards their original values before fine-tuning commenced.

\begin{figure*}[ht]
    \centering
    \begin{tabular}[c]{@{}c@{}}
    \begin{subfigure}[c]{0.58\linewidth}
        \centering
        \includegraphics[width=\linewidth]{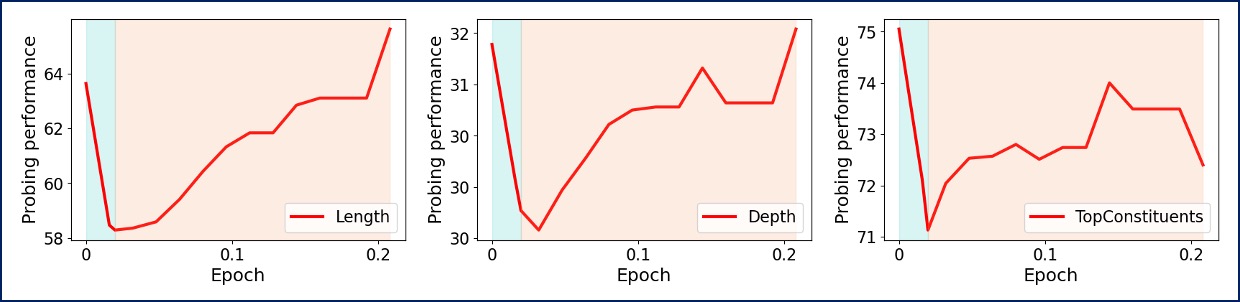}
        \caption{Group1: Tasks showing deterioration in Phase 1 followed by recovery in Phase 2.}
        \label{fig:probing_all (a)}
    \end{subfigure} \\
    \noalign{\bigskip}
    \begin{subfigure}[c]{0.58\linewidth}
        \includegraphics[width=\linewidth]{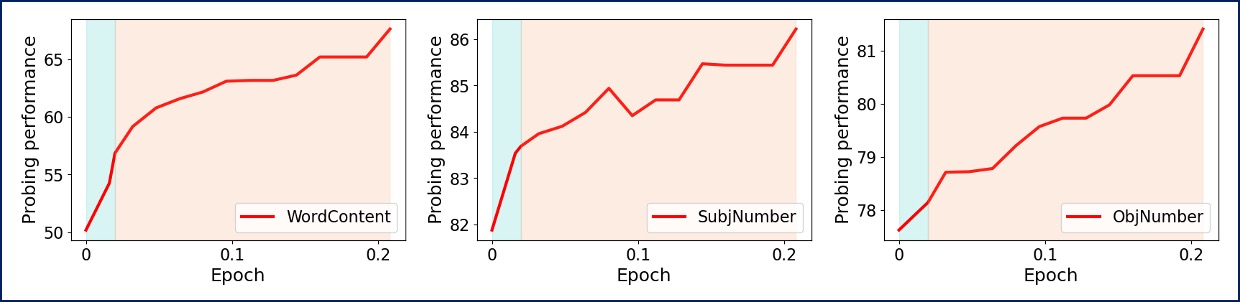}
        \caption{Group2: Tasks showing a continuous improvement.}
        \label{fig:probing_all (b)}
    \end{subfigure}
    \end{tabular}
    \hfill
    \begin{subfigure}[c]{0.38\linewidth}
        \includegraphics[width=\linewidth]{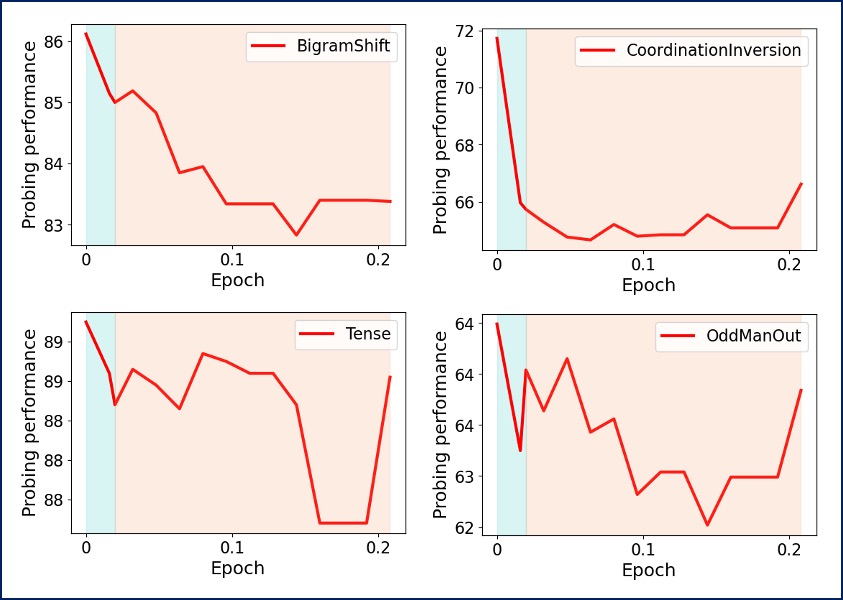}
        \caption{Group3: Tasks showing a decline or oscillation.}
        \label{fig:probing_all (c)}
    \end{subfigure}
    \caption{Training dynamics of ten linguistic abilities. We categorize the ten tasks into three groups based on the trends in probing performance.}
    \label{fig:probing_all}
\end{figure*}

\subsection{Linguistic Abilities}
\label{subsec:linguistic_abilities}

The purpose of fine-tuning a language model is to leverage the language model's pre-learned linguistic abilities. However, due to the use of cosine similarity as the measure of sentence embedding, fine-tuning with contrastive loss becomes imperative. For the case of BERT, it becomes essential to enhance uniformity through fine-tuning despite the possibility of negatively affecting BERT's linguistic abilities. 
To explore linguistic aspects, we followed the methodology of~\citet{conneau2018you} and utilized SentEval toolkit\footnote{\url{https://github.com/facebookresearch/SentEval/tree/main/data/probing}} to investigate training dynamics of ten different linguistic abilities. Based on the observed trends in training dynamics, we grouped the ten probing tasks into three categories, as presented in Figure~\ref{fig:probing_all}.

The first group (Figure~\ref{fig:probing_all (a)}) consists of three tasks that exhibit deterioration in Phase 1 followed by recovery in Phase 2.
They are Length~(number of tokens), Depth~(depth of sentence structure trees), and TopConstituents~(the grammatical structure of sentences), and all three are closely related to the performance of sentence embedding. 
We emphasize that the worst performance occurs at or near the boundary between Phase 1 and Phase 2 for the three linguistic abilities, indicating a strong correlation with representation rank. The trend of deterioration followed by recovery was also observed for alignment, where the deterioration occurs while uniformity sharply improves.

The second group (Figure~\ref{fig:probing_all (b)}) consists of three tasks that 
exhibit an upward performance trend in both Phase 1 and Phase 2. They are 
WordContent (deducing words from sentence representations), SubjNumber, and ObjNumber (matching the number of subjects and objects in sentence clauses, respectively), and all three are intimately related to the sentence embedding task. These three linguistic abilities do not deteriorate in Phase 1 despite uniformity's sharp improvement, suggesting that the three do not form a trade-off relationship with uniformity. 

The third group (Figure~\ref{fig:probing_all (c)}) consists of the four remaining tasks. Their performance either deteriorates (BigramShift and CoordinationInversion) or oscillates (Tense and OddManOut) throughout the fine-tuning.
Compared to the other six linguistic abilities, these four are less directly associated with the sentence embedding task. For instance, changes in the order of words or clauses within a sentence are likely to have a lesser impact on the semantic information of a sentence in most common use cases of sentence embedding.

\subsection{Extreme Correlation between Performance and Rank}
\label{subsec:performance_rank}
\begin{figure}[ht]
    \centering
    \begin{subfigure}[t]{0.23\textwidth}
        \includegraphics[width=\textwidth]{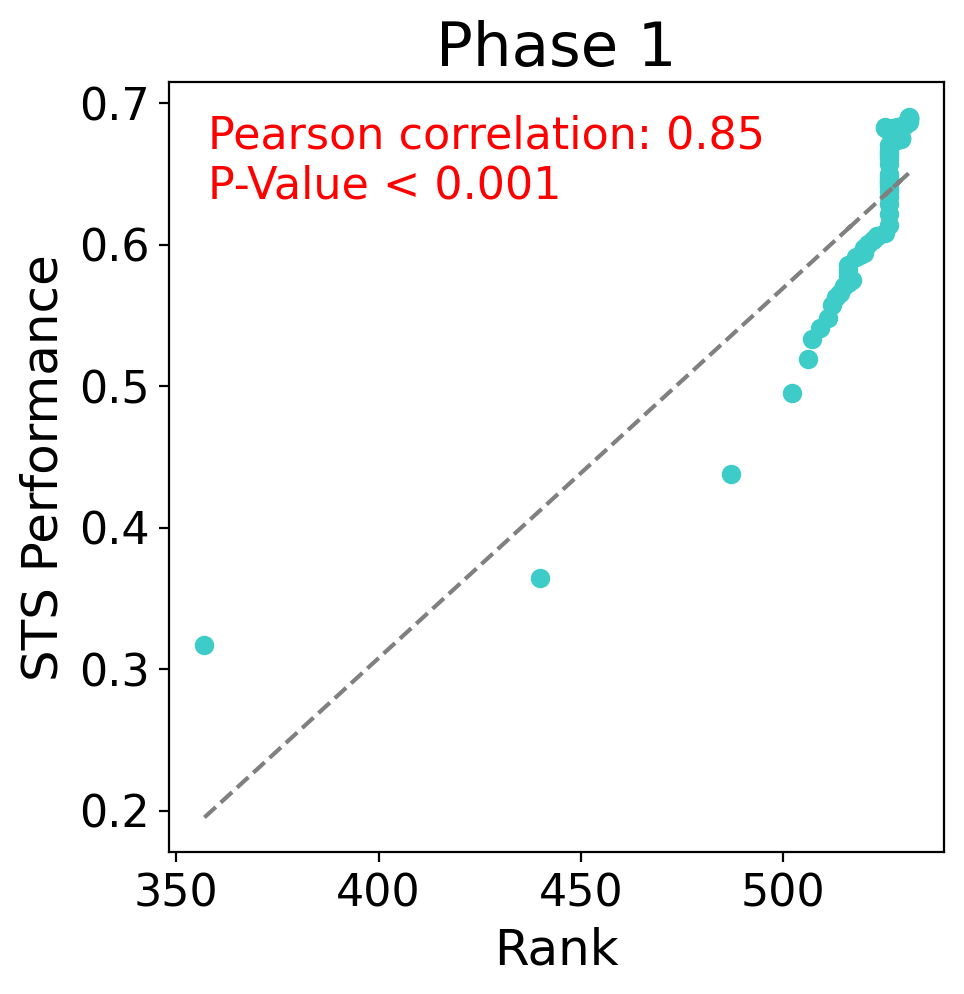}
        \caption{Scatter plot for Phase 1}
        \label{fig:relationship_rank_performance (a)}
    \end{subfigure}
    \hfill
    \begin{subfigure}[t]{0.23\textwidth}
        \includegraphics[width=\textwidth]{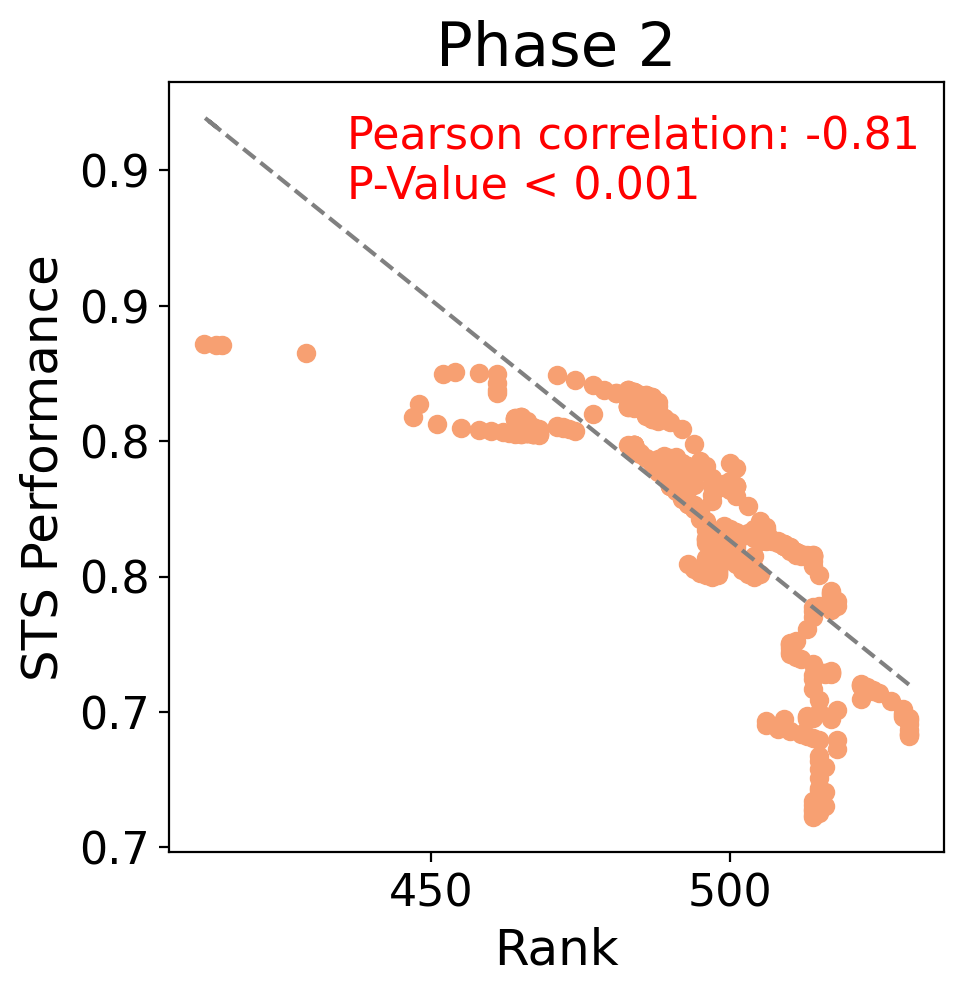}
        \caption{Scatter plot for Phase 2}
        \label{fig:relationship_rank_performance (b)}
    \end{subfigure}
    \caption{Scatter plot of rank and STS performance in two Phases. 
    We experimented with SimCSE-BERT$_\texttt{base}$ and recorded validation performance and rank every 5 steps during the training. The Pearson correlation is $0.85$ in Phase 1 and $-0.81$ in Phase 2.
    }
    \label{fig:relationship_rank_performance}
\end{figure}
In Sections~\ref{subsec:alignment_uniformity} and \ref{subsec:linguistic_abilities}, we explored the training dynamics of alignment and uniformity, as well as linguistic abilities, and their association with representation rank through the definition of Phase 1 and Phase 2. In this section, we demonstrate that representation rank is also closely linked to sentence embedding performance.

The scatter plots illustrating the relationship between representation rank and STS performance are presented in Figure~\ref{fig:relationship_rank_performance} for both Phase 1 and Phase 2. In Phase 1, the correlation between rank and performance is notably strong at $0.85$. This indicates that performance improves as rank increases. In Phase 2, the correlation remains strong but in a reversed manner at $-0.81$. This suggests that performance improves as rank decreases. The robust correlation values highlight the potential for enhancing sentence embedding by controlling representation rank, a possibility that we further investigate in the following section.

\section{Experiments -- Rank Reduction}
\label{sec:Experiments}

In this section, we introduce a rank regularization strategy that is straightforward to implement. Using this strategy, we demonstrate its ability to consistently enhance both the performance and stability of five state-of-the-art sentence embedding methods.
Details about the experimental setup and hyperparameter information are available in Appendix~\ref{sec:appendix_setup} and \ref{sec:appendix_HP}.

\subsection{Regularization of Representation Rank}
Algebraic rank is not differentiable because it is discrete. Instead, effective rank~\citep{roy2007effective} is known as a continuous-value extension that is differentiable. For the implementation of rank regularizer in our work, we follow the practical design provided in~\citet{Kim_2023_CVPR} where the exponential function part is dropped from effective rank. The regularization applied to SimCSE loss can be formulated as below. 

\begin{align}
    \mathcal{L}^{'}_\mathsf{SimCSE} &= \mathcal{L}_\mathsf{SimCSE} + \gamma \cdot \sum_j \lambda_{j} \log{\lambda_{j}}.
    \label{eq:RR_loss}
\end{align}

\subsection{Analysis of Phase-Dependency}
For the two phases identified in Section~\ref{sec:SimCSEandRank}, we investigate the effect of representation rank regularization. Specifically, we explore phase-dependent control over STS-B development set. The investigation results are summarized in Table~\ref{tab:phase_sepeartion_result}. 
In the table, the single strongest trend we can observe is that promoting rank reduction in Phase 2 is significantly helpful. As evidenced by the exemplary plots provided in Figure~\ref{fig:appendix_logging_3seeds} of Appendix~\ref{appendix:C}, rank reduction during Phase 2 facilitates rapid and stable fine-tuning, resulting in the attainment of best performance swiftly and consistently regardless of the chosen random seed. Apparently, rapid and stable fine-tuning is helpful for improving performance where rank reduction in Phase 2 promotes such a behavior. 
For Phase 1, it is unclear if either rank increase or rank decrease is helpful. This can be because of the steep change in uniformity during Phase 1, where rank regularizing cannot make a steady impact. 
Overall, the best performance is achieved when rank is increased in Phase 1 and decreased in Phase 2. Compared to the baseline, the average performance of SimCSE is significantly improved from 80.90 to 84.19. The best performance, however, is marginally better than 84.13 that is obtained by decreasing rank in both Phase 1 and Phase 2. 

Based on the aforementioned analysis, we opt for our rank strategy, which involves reducing rank in both Phase 1 and Phase 2. This strategy simplifies the application of rank regularization as there is no need to distinguish between Phase 1 and Phase 2. Such a distinction necessitates continuous monitoring of rank during fine-tuning, which consequently incurs a computational burden.
We name our strategy as RR~(Rank Reduction).

\begin{table}[t]
\centering
\small
\begin{tabular}{@{}rrcc@{}}
\toprule
\multicolumn{2}{c}{\textbf{Rank}} & \multicolumn{2}{c}{\textbf{Performance}} \\ \cmidrule(l){1-4}
\multicolumn{1}{c}{\textbf{Phase 1}} & \multicolumn{1}{c}{\textbf{Phase 2}} & \textbf{Avg.} ($\uparrow$) & \textbf{Std.}  ($\downarrow$) \\
\midrule \midrule
-~~~~~~~$(\gamma=0)$ & -~~~~~~~$(\gamma=0)$ & 80.90 & 1.40 \\
Increase~~$(\gamma<0)$ & -~~~~~~~$(\gamma=0)$ & 80.78 & \textbf{1.07} \\
Decrease~$(\gamma>0)$ & -~~~~~~~$(\gamma=0)$ & \textbf{81.35} & 1.28\\
\midrule
-~~~~~~~$(\gamma=0)$ & Increase~~$(\gamma<0)$ & 75.68 & 1.94 \\
Increase~~$(\gamma<0)$ & Increase~~$(\gamma<0)$ & 75.69 & 1.87 \\
Decrease~$(\gamma>0)$ & Increase~~$(\gamma<0)$ & \textbf{76.70} & \textbf{1.65} \\
\midrule
-~~~~~~~$(\gamma=0)$ & Decrease~$(\gamma>0)$ & 84.03 & 0.42 \\
Increase~~$(\gamma<0)$ & Decrease~$(\gamma>0)$ & \textbf{84.19} & \textbf{0.33} \\
Decrease~$(\gamma>0)$ & Decrease~$(\gamma>0)$ & 84.13 & 0.44 \\
\bottomrule

\end{tabular}
\caption{
The effect of phase-dependent rank regularization. 
The performance of fine-tuned SimCSE-BERT$_\texttt{base}$ was evaluated with STS-B development set. The evaluation was repeated 10 times with different random seeds. The average correlation performance in percentage and the standard deviation are shown.
}
\label{tab:phase_sepeartion_result}
\end{table}

\begin{table*}[!t]
\centering
\resizebox{\textwidth}{!}{
\begin{tabular}{@{}lcccccccll@{}}
\toprule
\textbf{Model} & \multicolumn{8}{c}{\textbf{STS Performance}} & \multicolumn{1}{c}{\textbf{Rank}}\\ \cmidrule(l){2-9} \cmidrule(l){10-10}
& \multicolumn{1}{c}{\textbf{STS12}} & \multicolumn{1}{c}{\textbf{STS13}} & \multicolumn{1}{c}{\textbf{STS14}} & \multicolumn{1}{c}{\textbf{STS15}} & \multicolumn{1}{c}{\textbf{STS16}} & \multicolumn{1}{c}{\textbf{STS-B}} & \multicolumn{1}{c}{\textbf{SICK-R}} & \multicolumn{1}{c}{\textbf{Avg.}} & \multicolumn{1}{c}{\textbf{Avg.}} \\ \midrule \midrule
SimCSE$^\dag$ & 68.40 & 82.41 & 74.38 & 80.91 & 78.56 & 76.85 & 72.23 & 76.25 & 324 \\
\ \ \ \ \ with RR & 71.36  & 83.56 & 75.66 & 83.50 & 80.14 & 79.94 & 72.03 & 78.03 (+1.78) & 195 (-129) \\
\midrule
MixCSE & 70.55 & 82.85 & 75.81 & 83.26 & 79.34 & 79.23 & 72.21 & 77.61 & 246 \\
\ \ \ \ \ with RR & 70.56 & 82.81 & 76.17 & 83.07 & 79.50 & 79.70 & 73.46 & 77.90 (+0.29) & 197 (-49) \\
\midrule
ESimCSE$^\ddag$ & 73.40  & 83.27 & 77.25 & 82.66 & 78.81 & 80.17 & 72.30  & 78.26 & 246 \\
\ \ \ \ \ with RR & 73.01 & 84.57 & 77.09 & 83.60 & 78.81 & 79.90 & 71.88 & 78.41 (+0.15) & 230 (-16) \\
\midrule
InfoCSE$^\S$ & 70.53 & 84.59 & 76.40 & 85.10 & 81.95 & 82.00 & 71.37 & 78.85 & 202 \\
\ \ \ \ \ with RR & 72.79 & 84.8  & 75.83 & 84.36 & 81.40 & 81.72 & 71.66 & 78.94 (+0.09) & 190 (-11) \\
\midrule
PromptBERT$^\P$ & 71.98 & 84.66 & 77.13 & 84.52 & 81.10 & 82.02 & 70.64 & 78.86 & 219
\\
\ \ \ \ \ with RR$^\clubsuit$ & 73.96 & 84.78 & 77.56 & 84.11 & 79.68 & 82.04 & 70.37 & 78.93 (+0.07) & 198 (-21) \\
\bottomrule
\end{tabular}
}
\caption{Performance and rank of five CL-based models. Each of these models employs \texttt{bert-base-uncased} as its backbone. For all five models, RR~(Rank Reduction) consistently improves the average performance over the 7 datasets. 
\dag: results from (\citet{gao2021simcse}), \ddag: results from (\citet{wu2021esimcse}), \S: results from (\citet{wu2022infocse}), \P: results from (\citet{jiang2022promptbert}).
Results for MixCSE-BERT were reproduced because pre-trained weights were not publicly available. $\clubsuit$: results from Table~\ref{tab:promptbert_denoising} where denoising is replaced by RR.
}
\label{tab:STS_performance}
\end{table*}

\subsection{Performance Improvement}
To investigate the effectiveness of RR, we have applied RR to five CL-based sentence embedding models. The results are shown in Table~\ref{tab:STS_performance}. Performance improvement by RR can be observed for all five models and the visualization of the improvement can be found in Figure~\ref{fig:rank_performance}. Except for ESimCSE, the final rank with RR regularization ends up in a narrow range between 190 and 198. Also, it is interesting to note that the performance gap between state-of-the-art PromptBERT and SimCSE is substantially reduced from 2.61~($=78.86-76.25$) to 0.90~($=78.93-78.03$) after applying RR. 

In addition to the results in Table~\ref{tab:STS_performance}, we provide two supplementary results in Table~\ref{tab:STS_performance_various_bert} of Appendix~\ref{sec:appendix_various_bert}. Firstly, we investigate dimensionality dependency by examining three BERT models with hidden state dimensionality of 512, 768, and 1024. Secondly, we investigate two RoBERTa models with dimensionality of 768 and 1024. In all experiments, we consistently observed a performance improvement with RR. The magnitude of improvement was contingent upon the performance of the original model. For models with high initial performance, the improvement was relatively less pronounced. This phenomenon is likely attributed to these models nearing the saturation point of STS performance.

\begin{figure}[t]
    \centering
    \includegraphics[width=0.9\columnwidth]{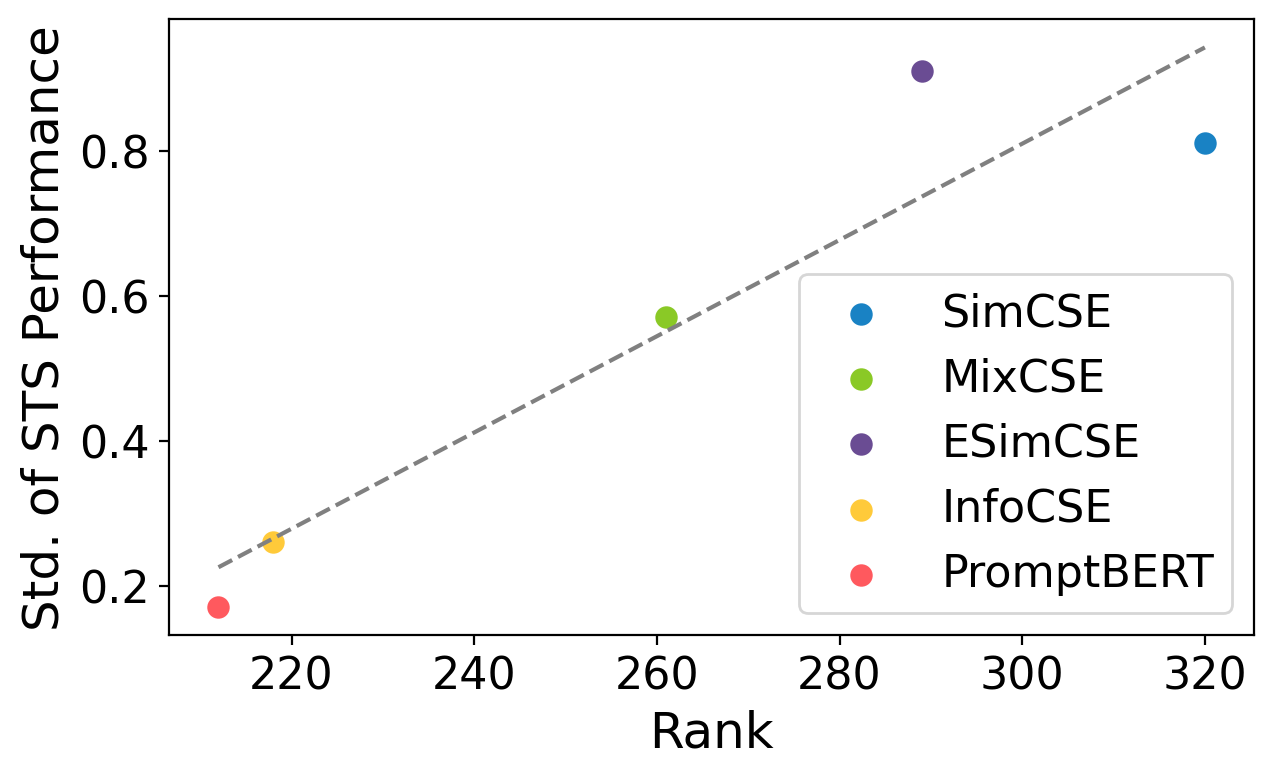}
    \caption{Rank vs. instability for five CL-based sentence embedding models. STS performance was assessed across 10 different random seeds, followed by the evaluation of their standard deviation of performance and average rank.}
    \label{fig:performance_stability_all_models}
\end{figure}

\begin{table}[t]
\centering
\small
\begin{tabular}{@{}lccc@{}}
\toprule
\multicolumn{1}{l}{\textbf{Std.}} & \multicolumn{2}{c}{\textbf{Model}} & 
\multicolumn{1}{l}{\textbf{Diff.}}\\ \cmidrule(l){2-3}
& \multicolumn{1}{c}{\textbf{Original}} & \multicolumn{1}{c}{\textbf{With RR}} &  \\ 
\midrule \midrule
SimCSE & 0.81 & 0.30 & -0.51\\
MixCSE & 0.57 &  0.50 & -0.07\\
ESimCSE & 0.91 & 0.73 & -0.18 \\
InfoCSE & 0.26 & 0.48 & +0.22 \\
PromptBERT & 0.17 & 0.17 & -0.00 \\
\bottomrule
\end{tabular}
\caption{
Effect of RR for improving stability. STS performance was assessed across 10 different random seeds, followed by the evaluation of their standard deviation. For InfoCSE and PromptBERT, the original models are already stable and RR does not improve their stability.
}
\label{tab:performance_stability}
\end{table}

\subsection{Stability Improvement}
Recent studies~\cite{jiang2022promptbert,zhang2022unsupervised} 
have highlighted an instability concern observed during the training of unsupervised SimCSE models. This concern is characterized by a significant variability in SimCSE performance depending on the random seed employed. In particular, learning methods that demonstrate instability, even if they achieve high performance for a small subset of selectively chosen seeds, are expected to show inferior overall performance.

The instability can also be demonstrated to have a strong correlation with representation rank. In Figure~\ref{fig:performance_stability_all_models}, we are showing a scatter plot between rank and instability for the five CL-based models. The Pearson correlation between the average rank and standard deviation is remarkably high at 0.94, indicating a clear and strong connection.
Therefore, adopting RR can be expected to enhance stability. The results of applying RR are shown in Table~\ref{tab:performance_stability}. For the three original models with standard deviation larger than 0.50, RR is effective in improving stability. However, for InfoCSE and PromptBERT, stability either deteriorates or remains consistent, likely owing to the already robust stability of the original models.

\section{Discussion}
\label{sec:discussion}

\subsection{Justification of Rank Regularization}
In Section~\ref{sec:SimCSEandRank}, we have explored key aspects of SimCSE, which is a foundational CL-based sentence embedding method. Among the various aspects that are significantly adjusted during fine-tuning, we chose rank as the target for regularization. This choice is superior to other options for the following reasons. 

Alignment and uniformity were investigated in Section~\ref{subsec:alignment_uniformity}. Because these concepts are directly related to the contrastive loss of SimCSE, they naturally improve without any modification in the SimCSE loss. Nevertheless, we attempted to further enhance alignment by applying an alignment regularizer. While this approach helped improve the performance of SimCSE, the improvement was approximately 1\% smaller than when using RR. This observation can be explained by the findings in~\citet{Kim_2023_CVPR}, where the implementational effectiveness of the loss function we adopt in Equation~\ref{eq:approx_rank} is demonstrated.

Linguistic abilities were explored in Section~\ref{subsec:linguistic_abilities}. It was discovered that certain probing tasks are directly related to the performance of sentence embedding. Consequently, leveraging probing tasks to enhance SimCSE appears logical. However, utilizing probing tasks necessitates human labeling, which contradicts the objective of unsupervised learning. Additionally, employing probing tasks carries a degree of risk, as it has been reported that different encoder architectures trained with the same objective can produce linguistically distinct embeddings~\citep{conneau2018you}.

\subsection{Rank Analysis of PromptBERT}
\label{subsec:discussion_analysis_PromptBERT}
PromptBERT~\citep{jiang2022promptbert} is recognized as a state-of-the-art method for sentence embedding. Here, we delve into its two essential concepts.
The first concept pertains to the selection of the prompt. To gain deeper insights into the impact of the prompt on performance and rank, we conducted a straightforward experiment by testing four different prompts. 
The results are presented in Table~\ref{tab:various prompts}. 
As anticipated, employing a simple prompt or a completely random prompt does not lead to a high performance. Surprisingly, however, employing a negated prompt achieves a performance level comparable to that of PromptBERT. This observation is counter-intuitive and suggests that the semantic meaning of the designed prompt alone cannot fully account for the performance gain of PromptBERT. In contrast, the representation rank consistently correlates with performance across the four cases, as indicated in the last column.

The second concept concerns template denoising. We compare the performance and representation rank when denoising and Rank Reduction (RR) are applied or not during training. As evident from Table~\ref{tab:promptbert_denoising}, removing denoising from PromptBERT results in a decline in sentence embedding performance and a substantial increase in representation rank. Introducing RR mitigates this increase in rank and leads to improved performance. Notably, the performance improvement achieved by excluding denoising and incorporating RR is more significant than applying RR in conjunction with denoising. This suggests a potential conflict in roles, indicating that denoising may interfere with the role of rank reduction in lowering representation rank.

\begin{table}[t] 
\centering
\resizebox{\columnwidth}{!}{
\begin{tabular}{@{}lllcc@{}}
\toprule
\multicolumn{2}{l}{\textbf{Prompt}} & \multicolumn{1}{l}{\textbf{Perf.}} & \multicolumn{1}{l} {\textbf{Rank}}\\ \cmidrule(l){0-1}
\multicolumn{1}{l}{\textbf{Case}} & \multicolumn{1}{l}{\textbf{Template}}& &\\

\midrule \midrule
\multirow{2}{*}{Simple}&{[}X{]} {[}MASK{]} . &  \multirow{2}{*}{76.96} & \multirow{2}{*}{263} \\
&{[}X{]} {[}MASK{]} . \\ \hline
\multirow{2}{*}{Random}&Apple banana “{[}X{]}” cherry {[}MASK{]} . &  \multirow{2}{*}{77.50} & \multirow{2}{*}{250} \\
&Apple banana “{[}X{]}” cherry {[}MASK{]} . \\ \hline
\multirow{2}{*}{Negated}&This sentence of “{[}X{]}” means {[}MASK{]} . &  \multirow{2}{*}{78.23} & \multirow{2}{*}{212} \\
&This sentence : “{[}X{]}” \textbf{does not} mean {[}MASK{]} . \\ \hline
\multirow{2}{*}{PromptBERT}&This sentence of “{[}X{]}” means {[}MASK{]} . &  \multirow{2}{*}{78.86$^\dag$} & \multirow{2}{*}{219} \\
&This sentence : “{[}X{]}” means {[}MASK{]} . \\\bottomrule
\end{tabular}
}
\caption{The effect of prompt modification for PromptBERT. 
PromptBERT was trained using four different prompts: a simple prompt, a prompt featuring unrelated words, a prompt incorporating negation, and the optimal prompt reported in PromptBERT paper.
\dag: results from (\citet{jiang2022promptbert}).
}
\label{tab:various prompts}
\end{table}

\begin{table}[!t]
    \centering
    \small
    \resizebox{\columnwidth}{!}{
    \begin{tabular}{@{}lcccc@{}}
        \toprule
        \multicolumn{1}{l}{\textbf{Model}} & \multicolumn{1}{l}{\textbf{Denoising}} & \multicolumn{1}{l}{\textbf{RR}} & \multicolumn{1}{l}{\textbf{Perf.}} & \multicolumn{1}{l}{\textbf{Rank}}\\
        \midrule \midrule
        PromptBERT & O & X & 78.86$^{\dag}$ & 219 \\
        \ \ \ \ \ $-$ denoising & X & X & 78.61 & 298 \\
        \ \ \ \ \ $+$ RR & O & O & 78.84 & 199 \\
        \ \ \ \ \ $-$ denoising $+$ RR & X & O & \textbf{78.93} & 198 \\
        \bottomrule
    \end{tabular}
    }
    \caption{The effects of denoising and Rank Reduction (RR) on the performance and representation rank of PromptBERT.
    \dag: results from (\citet{jiang2022promptbert}).}
    \label{tab:promptbert_denoising}
    \vspace{-0.5cm}
\end{table}

\section{Conclusion}
The latest advancements in sentence embedding methods typically leverage contrastive learning as the fine-tuning method. Through our analysis of various key aspects of these CL-based models, we have shown that representation rank can play a pivotal role not only in analysis but also in regularization of the models.

\section{Limitations}
In this study, we did not provide a theoretical explanation for why CL-based models perform better at lower ranks, which remains a limitation of our research. Hence, it would be beneficial for future investigations to explore the theoretical relationship between rank and performance. Furthermore, given that the primary objective of this study is to conduct an analysis of prior works and enhance them accordingly, our study is not susceptible to potential risks.

\section*{Acknowledgement}
\label{sec:acknowledge}
This work was supported by a National Research Foundation of Korea (NRF) grant funded by the Korea government (MSIT) (No. NRF-2020R1A2C2007139) and in part by Institute of Information \& Communications Technology Planning \& Evaluation (IITP) grant funded by the Korea government(MSIT) [NO.2021-0-01343, Artificial Intelligence Graduate School Program (Seoul National University)] and Basic Science Research Program through the National Research Foundation of Korea(NRF) funded by the Ministry of Education(NRF-2022R1A6A1A03063039).

\bibliography{acl2023}

\clearpage
\appendix
\section{Experimental Setup}
\label{sec:appendix_setup}
For our experiments on five CL-based models, we referenced the official GitHub repository of the original paper and conducted our experiments on RTX3090 GPUs (24GB).
To reproduce the STS performance and compute the rank of the original models, we utilized model checkpoints published on Hugging Face to compute the performance and rank of the original models, excluding MixCSE.

We followed the settings of SimCSE, specifically its unsupervised variant.
We employed the training and testing datasets commonly used for unsupervised CL-based models. 
The models were trained on a dataset with one million sentences from Wikipedia. For the evaluation of models on the semantic textual similarity (STS) tasks, we employed SentEval toolkit~\citep{conneau2018senteval} and datasets of STS 2012-16~\citep{agirre2012semeval,agirre2013sem,agirre2014semeval,agirre2015semeval,agirre2016semeval}, STS-B~\citep{cer2017semeval}, and SICK-R~\citep{marelli2014sick}.

\section{Hyperparameters}
\label{sec:appendix_HP}
To fine-tune SimCSE models based on BERT~\citep{devlin2018bert} and RoBERTa~\citep{liu2019roberta}, we used the hyperparameter settings reported as optimal by~\citet{gao2021simcse}. 
When training the models with the RR (Eq.~\ref{eq:RR_loss}) loss, we adjusted the learning rate $\in\{$1e-5, 3e-5, 5e-5$\}$ following the convention of the SimCSE paper. On a single RTX3090 GPU, SimCSE-BERT$_\texttt{base}$ with RR completed training in approximately one hour.

For other CL-based models, except for PromptBERT, we used the best learning rate suggested by the original papers.
To determine the best RR loss coefficient, we carried out grid-search of regularizer coefficient $\in\{$5e-1, 1e-1, 5e-2, 1e-2, 8e-3, 6e-3, 4e-3, 2e-3, 1e-3$\}$. For cases where the rank of representations is already low, such as InfoCSE or PromptBERT, additional coefficients such as 1e-5 and 5e-5 were explored.
After selecting the coefficient, we derived the final results using ten different random seeds.

The hyperparameter settings employed for fine-tuning SimCSE, MixCSE, ESimCSE, InfoCSE, and PromptBERT are detailed in Table~\ref{tab:appendix_simcse_hp} and \ref{tab:appendix_other_models_hp}, including learning rate, batch size, and coefficient for the RR regularizer here. For other hyperparameters not explicitly mentioned, we followed the settings presented in the original papers.

\begin{table}[ht]
\centering
\small
\begin{tabular}{lccccc}
\toprule
\multicolumn{1}{c}{\multirow{2}{*}{\textbf{Model}}} & \multicolumn{1}{c}{\multirow{2}{*}{\textbf{HP}}}  & \multicolumn{2}{c}{\textbf{BERT}} & \multicolumn{2}{c}{\textbf{RoBERTa}} \\ \cmidrule(l){3-4} \cmidrule(l){5-6}
 & & \textbf{base} & \textbf{large}  & \textbf{base} & \textbf{large} \\
\midrule \midrule
\multirow{2}{*}{SimCSE} & bs & 64 & 64 & 512 & 512\\
& lr & 3e-5 & 1e-5 & 1e-5 & 3e-5\\
\midrule
\multirow{3}{*}{\shortstack{SimCSE\\w/ RR}} & bs & 64 & 64 & 512 & 512\\
& lr & 1e-5 & 1e-5 & 1e-5 & 3e-5\\
& RR coef & 1e-3 & 2e-3 & 1e-3 & 1e-3 \\
\bottomrule
\end{tabular}
\caption{Hyperparameters for training SimCSE. We present hyperparameters (HP) such as batch size (bs), learning rate (lr), and coefficient for the RR regularizer (RR coef).}
\label{tab:appendix_simcse_hp}
\end{table}

\begin{table}[ht]
\centering
\small
\begin{tabular}{lccc}
\toprule
\multicolumn{1}{c}{\multirow{2}{*}{\textbf{Model}}} & \multicolumn{3}{c}{\textbf{BERT$_\texttt{base}$}}\\ \cmidrule(l){2-4}
& \textbf{bs} & \textbf{lr} & \textbf{RR coef} \\
\midrule \midrule
MixCSE & 64 & 3e-5 & - \\
\ \ \ \ \ w/ RR & 64 & 3e-5 & 1e-3\\
\midrule
ESimCSE & 64 & 3e-5 & - \\
\ \ \ \ \ w/ RR & 64 & 3e-5 & 1e-3 \\
\midrule
InfoCSE & 64 & 7e-6 & - \\
\ \ \ \ \ w/ RR & 64 & 7e-6 & 1e-4 \\
\midrule
PromptBERT & 256 & 1e-5 & - \\
\ \ \ \ \ wo/ denoising & 256 & 1e-5 & - \\
\ \ \ \ \ w/ RR & 256 & 1e-5 & 5e-4 \\
\ \ \ \ \ wo/ denoising w/ RR & 256 & 3e-6 & 3e-3 \\
\bottomrule
\end{tabular}
\caption{Hyperparameters for training models including MixCSE, ESimCSE, InfoCSE, and PromptBERT. We present hyperparameters such as batch size (bs), learning rate (lr), and coefficient for the RR regularizer (RR coef). For other hyperparameters, we followed the configurations suggested by the original papers.}
\label{tab:appendix_other_models_hp}
\end{table}

\section{Comparison of Phase 2 Endpoints of SimCSE and SimCSE with RR}
\label{appendix:C}
\begin{figure*}[ht]
\begin{subfigure}[b]{0.45\textwidth}
    \includegraphics[width=\textwidth]{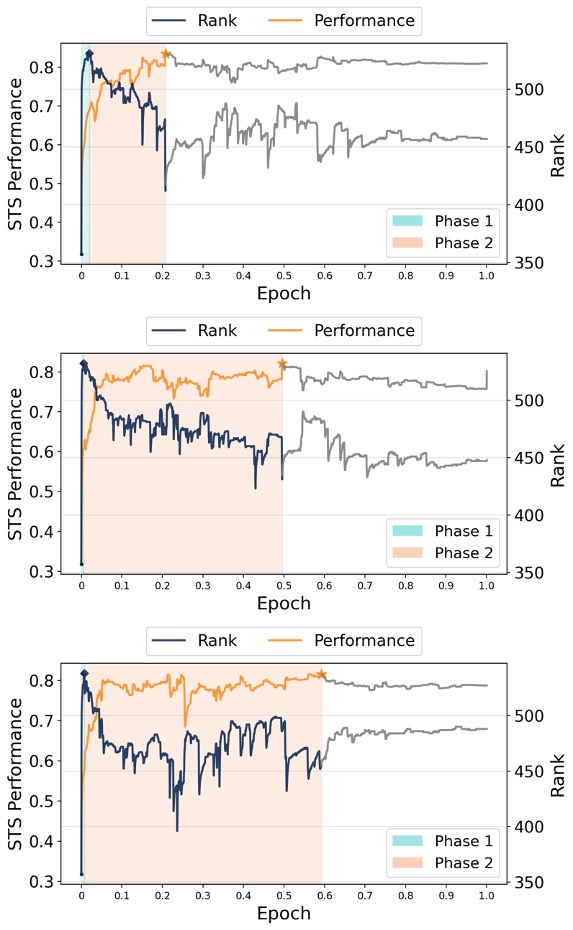}
    \caption{SimCSE}
    \label{fig:appendix_logging_3seeds (a)}
\end{subfigure}
\hfill
\begin{subfigure}[b]{0.45\textwidth}
    \includegraphics[width=\textwidth]{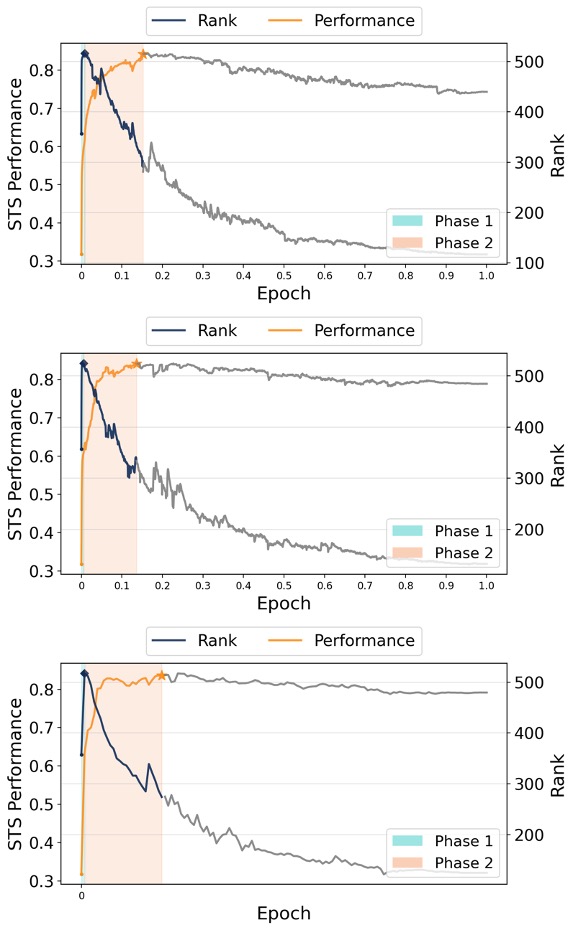}
    \caption{SimCSE with RR}
    \label{fig:appendix_logging_3seeds (b)}
\end{subfigure}
    \caption{Phase 2 endpoints of SimCSE and SimCSE with RR. Applying RR results in endpoints occurring more rapidly and consistently.}
    \label{fig:appendix_logging_3seeds} 
\end{figure*}

In Figure~\ref{fig:appendix_logging_3seeds}, we compare Phase 2 endpoints of the original SimCSE and SimCSE with RR, each using three different random seeds. The comparison reveals noticeable variability in the Phase 2 endpoints of the original SimCSE models (Figure~\ref{fig:appendix_logging_3seeds (a)}) as opposed to those models with RR (Figure~\ref{fig:appendix_logging_3seeds (b)}). The introduction of RR regularization not only contributes to stabilizing the point of early stopping but also expedites the training process. 
This enables shortening of fine-tuning, which might contribute towards preserving linguistic abilities of the pre-trained language model.

\section{Rank and STS Performance of BERT and RoBERTa Models}
\label{sec:appendix_various_bert}
While the primary focus of the main manuscript is on the outcomes for SimCSE models employing BERT-base configuration, this section extends the analysis to various configurations of BERT and RoBERTa models (BERT-base, BERT-large, RoBERTa-base, and RoBERTa-large). 
Furthermore, to explore the impact of reduced dimensionality of hidden states on performance and representational rank, we conducted experiments using a model structurally analogous to BERT-base, albeit with a hidden state dimensionality reduced to 512.
As illustrated in Table~\ref{tab:STS_performance_various_bert}, the application of RR across all models results in a decrease in rank and an enhancement in STS performance. 
Additionally, we observe that an increase in the dimensionality of the language model correlates with improved performance and higher ranks of representations.
This suggests that models with greater computational capabilities tend to deliver superior STS performance, and that diminishing the rank within a given dimensionality tends to boost performance.
It was also noted that the representational ranks of RoBERTa models are generally higher than those of BERT models, a phenomenon that may be ascribed to the variances in the datasets and pre-training methodologies employed in the development of BERT and RoBERTa.

\begin{table*}[ht]
\centering
\resizebox{\textwidth}{!}{
\begin{tabular}{@{}lccccccccll@{}}
\toprule
\multirow{2}{*}{\textbf{Model}} & \multirow{2}{*}{\textbf{Dim.}} & \multicolumn{8}{c}{\textbf{STS Performance}} & \multicolumn{1}{c}{\textbf{Rank}}\\ \cmidrule(l){3-10} \cmidrule(l){11-11}
& & \multicolumn{1}{c}{\textbf{STS12}} & \multicolumn{1}{c}{\textbf{STS13}} & \multicolumn{1}{c}{\textbf{STS14}} & \multicolumn{1}{c}{\textbf{STS15}} & \multicolumn{1}{c}{\textbf{STS16}} & \multicolumn{1}{c}{\textbf{STS-B}} & \multicolumn{1}{c}{\textbf{SICK-R}} & \multicolumn{1}{c}{\textbf{Avg.}} & \multicolumn{1}{c}{\textbf{Avg.}} \\ \midrule \midrule
SimCSE-BERT (512) & 512 & 65.98 & 78.49 & 71.63 & 80.25 & 76.98 & 75.56 & 66.57 & 73.64 & 251 \\
\ \ \ \ \ with RR & 512 & 67.68 & 79.32 & 72.58 & 81.23 & 77.85 & 78.19 & 68.51 & 75.05 (+1.41) & 212 (-39) \\
\midrule
SimCSE-BERT$_\texttt{base}$$^\dag$ & 768 & 68.40 & 82.41 & 74.38 & 80.91 & 78.56 & 76.85 & 72.23 & 76.25 & 324 \\
\ \ \ \ \ with RR & 768 & 71.36  & 83.56 & 75.66 & 83.50 & 80.14 & 79.94 & 72.03 & 78.03 (+1.78) & 195 (-129) \\
\midrule
SimCSE-BERT$_\texttt{large}$ & 1024 & 70.88 & 84.16 & 76.43 & 84.50 & 79.76 & 79.26 & 73.88 & 78.41 & 452 \\
\ \ \ \ \ with RR & 1024 & 72.74 & 84.38 & 76.68 & 84.41 & 79.76 & 80.45 & 74.01 & 78.92 (+0.51) & 248 (-204) \\
\midrule \midrule
SimCSE-RoBERTa$_\texttt{base}$$^\dag$ & 768 & 70.16 & 81.77 & 73.24 & 81.36 & 80.65 & 80.22 & 68.56 & 76.57 & 356 \\
\ \ \ \ \ with RR & 768 & 70.65  & 82.03 & 74.46 & 83.06 & 81.20 & 80.91 & 69.03 & 77.33 (+0.76) & 307 (-48) \\
\midrule
SimCSE-RoBERTa$_\texttt{large}$$^\dag$ & 1024 & 72.86 & 83.99 & 75.62 & 84.77 & 81.80 & 81.98 & 71.26 & 78.90 & 455 \\
\ \ \ \ \ with RR & 1024 & 73.45 & 84.63 & 75.70  & 84.91   & 81.18   & 82.02  & 71.83  & 79.10 (+0.20) & 403 (-52) \\
\bottomrule
\end{tabular}
}
\caption{Performance and rank comparison of SimCSE models and models with RR regularizer on seven STS datasets. We provide the average rank values across the seven datasets for ease of comprehension. The term \textit{Dim.} denotes the dimensionality of the hidden states of language models.
In order to explore trends across language models, we conducted experiments with various configurations, including BERT-base and RoBERTa-base using 768 dimensions, as well as RoBERTa-base and BERT-large using 1024 dimensions.
Additionally, experiments were carried out with a BERT-512 version, which has the same configuration as BERT-base but with hidden states reduced to 512 dimensions. \dag: results from (\citet{gao2021simcse}). 
}
\label{tab:STS_performance_various_bert}
\end{table*}

\end{document}